\documentclass[a4paper,twoside]{article}

\usepackage{epsfig}
\usepackage{subfigure}
\usepackage{calc}
\usepackage{amssymb}
\usepackage{amstext}
\usepackage{amsmath}
\usepackage{amsthm}
\usepackage{multicol}
\usepackage{pslatex}
\usepackage{apalike}

\usepackage{cite}
\usepackage{graphicx}
\usepackage{csvsimple}
\usepackage[export]{adjustbox}
\usepackage{multirow}
\usepackage[table,xcdraw]{xcolor}
\usepackage{float}

\usepackage{SCITEPRESS}     

\begin{document}

\title{Detection of Human Rights Violations in Images: Can Convolutional Neural Networks help?}

\author{\authorname{Grigorios Kalliatakis\sup{1}, Shoaib Ehsan\sup{1}, Maria Fasli\sup{1}, Ales Leonardis\sup{2}, Juergen Gall\sup{3} and Klaus D. McDonald-Maier\sup{1}}
\affiliation{\sup{1}School of Computer Science and Electronic Engineering, University of Essex, UK}
\affiliation{\sup{2} School of Computer Science, University of Birmingham, UK}
\affiliation{\sup{3} Institute of Computer Science III, University of Bonn, Germany}
\email{\{gkallia, sehsan, mfasli, kdm\}@essex.ac.uk, a.leonardis@cs.bham.ac.uk, gall@iai.uni-bonn.de}
}


\abstract{After setting the performance benchmarks for image, video, speech and audio processing, deep convolutional networks have been core to the greatest advances in image recognition tasks in recent times.
This raises the question of whether there are any benefit in targeting these remarkable deep architectures with the unattempted task of recognising human rights violations through digital images. Under this perspective, we introduce a new, well-sampled human rights-centric dataset called Human Rights Understanding (HRUN). We conduct a rigorous evaluation on a common ground by combining this dataset with different state-of-the-art deep convolutional architectures in order to achieve recognition of human rights violations. Experimental results on the HRUN dataset have shown that the best performing CNN architectures can achieve up to 88.10\% mean average precision. Additionally, our experiments demonstrate that increasing the size of the training samples is crucial for achieving an improvement on mean average precision
principally when utilising very deep networks.}

\onecolumn \maketitle \normalsize \vfill

\section{\uppercase{Introduction}}
\label{sec:introduction}

\noindent Human rights violations continue to take place in many parts of the world today, while they have been ongoing during the entire human history. These days, organizations concerned with human rights are increasingly using digital images as a mechanism for supporting the exposure of human rights and international humanitarian law violations. However, utilising current advances in technology for studying, prosecuting and possibly preventing such misconduct from occurring have not yet made any progress. From this perspective, supporting human rights is seen as one scientific domain that could be strengthened by the latest developments in computer vision.
To support the continued growth of images and videos in human rights and international humanitarian law monitoring campaigns, this study examines how vision based systems can support human rights monitoring efforts by accurately detecting and identifying human rights violations utilising digital images. 

This work is made possible by recent progress in Convolutional Neural Networks (CNNs) \cite{1}, which has changed the landscape for well-studied computer vision tasks, such as image classification and object detection \cite{29, 30}, by comprehensively outperforming the initial handcrafted approaches \cite{17, 18, 3}. These state-of-the-art architectures are now finding their way into a number of vision based applications \cite{19, 20, 4}. 

\begin{figure*}[!t]
\centering
\includegraphics[height=6.5cm,width=13.25cm,keepaspectratio,center]{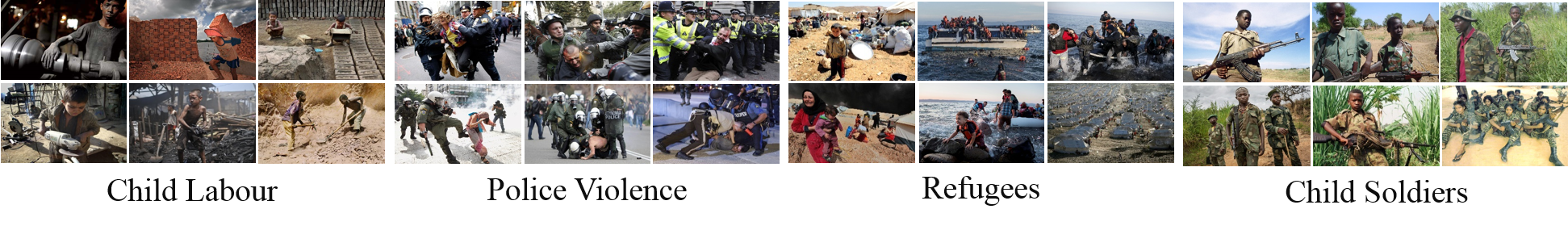}
\caption{Examples from all 4 categories of the Human Rights Understanding (HRUN) Dataset.}
\label{HRUN_examples}
\end{figure*}

A major contribution of our paper is a new, well-sampled human rights-centric dataset, called the \textit{Human Rights Understanding (HRUN)} dataset, which consists of 4 different categories of human rights violations and 100 diverse images per category. In this paper, we formulate the human rights violation recognition problem as being able to recognise a given input image (from the HRUN dataset) as belonging to one of these 4 categories of human rights violations. See Figure \ref{HRUN_examples} for examples of our data. We use this data for human rights understanding by evaluating different deep representations on this new dataset, while we perform experiments that illustrate the effect of network architecture, image context, and training data size on the accuracy of the system. 

In summary, our contribution is two-fold. Firstly we introduce a new human rights-centric dataset HRUN. Secondly, motivated by the great success of deep convolutional networks, we conduct a large set of rigorous experiments for the task of recognising human rights violations. As part of our tests, we delve into the latest, top-performing pre-trained deep convolutional models, allowing a fair, unbiased comparison on a common ground; something that has been largely missing so far in the literature. The remainder of the paper is organised as follows. Section 2 looks into prior works on database construction and image understanding with deep convolutional networks. Section 3 describes the methodology utilised for building our pioneer human rights understanding dataset. Section 4 demonstrates the classification pipeline used for the experiments, while the evaluation results are presented in Section 5, alongside a thorough discussion. Finally, conclusions and future directions are given in Section 6.

\section{\uppercase{Prior Work}}
\subsection{Database Construction}
\noindent Challenging databases are important for many areas of research, while large-scale datasets combined with CNNs have been key to recent advances in computer vision and machine learning applications. While the field of computer vision has developed several databases to organize knowledge about object categories \cite{8, 21, 22, 23}, scenes \cite{16, 24} or materials \cite{25, 26, 27} a well-inspected dataset of images depicting human rights violations does not currently exist.
The first reference point in standardized dataset of images and annotations was the VOC2010 dataset \cite{15}, which was constructed by utilizing images collected by non-vision/machine learning researchers, by querying Flickr with a number of related keywords, including the class name, synonyms and scenes or situations where the class is likely to appear. Similarly, an extensive scene understanding (SUN) database was introduced by \cite{24}, containing 899 environments and 130,519 images. The primary objectives of this work were to build the most complete dataset of scene image categories. Microsoft's work in regard to detection and segmentation of objects taking place in their natural context, was marked with the introduction of common objects in context \cite{33} (MS COCO) dataset including 328,000 images of complex everyday scenes consisted of 91 different object categories and 2.5 million labelled instances.
More recently \cite{28} presented their first version of a scene-centric database (LSUN) with millions of label images in each category alongside an integrated framework which makes use of deep learning techniques in order to achieve large-scale image annotation. To our knowledge, this particular work is the first attempt to construct a well-sampled image database in the domain of human rights understanding.

\subsection{Deep Convolutional Networks}
\noindent For decades, traditional machine learning systems demanded accurate engineering and significant domain expertise in order to design a feature extractor capable of converting raw data (such as the pixel values of an image) into a convenient internal representation or feature vector from which a classifier could classify or detect patterns in the input. Today, representation learning methods and principally CNNs \cite{1} are driving advances at a dramatic pace in the computer vision field after enjoying a great success in large-scale image recognition and object detection tasks \cite{2, 3, 4, 5, 6, 7}. The key aspect of deep learning representations is that the layers of features are not manually hand-crafted, but are learned from data using a generic-purpose learning scheme. The architecture of a typical deep-learning system can be considered as a multilayer stack of simple modules, each one transforming its input to increase both the selectivity and the invariance of the representation as stated in \cite{7}. In the last few years vision tasks became feasible due to high-performance computing systems such as GPUs, extensive public image repositories \cite{8}, a new regularisation technique called \textit{dropout} \cite{9} which prevents deep learning systems from overfitting, rectified linear units (ReLU) \cite{34}, softmax
layer and techniques able to generate more training examples by deforming the existing ones. 

Since \cite{2} first used an eight layer CNN (also known as AlexNet) trained on ImageNet to perform 1000-way object classification, a number of other works have used deep convolutional networks (ConvNets) to elevate image classification further \cite{10, 11, 12, 13, 14}. \cite{10} use a very deep CNN (also known as VGGNet) with up to 19 weight layers for large-scale image classification. They demonstrated that a substantially increased depth of a conventional ConvNet \cite{1, 2} can result in state-of-the-art performance on the ImageNet challenge dataset \cite{8}. They also perform localization for the same challenge by training a very deep ConvNet to predict the bounding box location instead of the class scores at the last fully connected layer. Another deep network architecture that has been recently used to great success is the GoogLeNet model of \cite{12} where an inception layer is composed of a shortcut branch and a few deeper branches in order to improve utilization of the computing resources inside the network. The two main ideas of that architecture are: (i) to create a multi-scale architecture capable of mirroring correlation structure in images and (ii) dimensional reduction and projections to keep their representation sparse along each spatial scale. Most recently \cite{11} announced the even deeper residual network (also known as ResNet), featured 152 layers, which has considerably improved the state-of-the-art performance of ImageNet \cite{8} classification and object detection on PASCAL \cite{15}. Residual networks are inspired by the observation that neural networks lean towards gaining higher training errors as the depth of the network increases to very large values. The authors argue that although the network gains more parameters by increasing its depth, the network becomes inferior at function approximation because of the gradients and training signals loss when they are propagated through numerous layers. Therefore, they give convincing theoretical and practical evidence that residual connections (reformulated layers for learning residual functions with reference to the layer input) are inherently necessary for training very deep convolutional models.

Outside of the aforementioned top-performing networks, other works worth mentioning are: \cite{14} where a rigorous evaluation study on different CNN architectures for the task of object recognition was conducted and \cite{16} where a brand-new scene-centric database called Places was introduced and established state-of-the-art results on different scene recognition tasks, by learning deep representations from their extensive database. Despite these impressive results, human rights advocacy is one of the high profile domains which remain broadly missing from the curated list of problems which were benefited from the continuing growth of deep convolutional networks. We build on this body of work in deep learning to solve the untrodden problem of recognising human rights violations utilising digital images.

\section{\uppercase{HRUN Dataset}}
\noindent Recent achievements in computer vision can be mainly ascribed to the ever growing size of visual knowledge in terms of labelled instances of objects, scenes, actions, attributes, and the dependent relationships between them. Therefore, obtaining effective high-level representations has become increasingly important, while a key question arises in the context of human rights understanding: \textit{how will we gather this structured visual knowledge?}

\begin{figure}[!t]
\centering
\includegraphics[height=7cm,width=7.50cm,keepaspectratio,center]{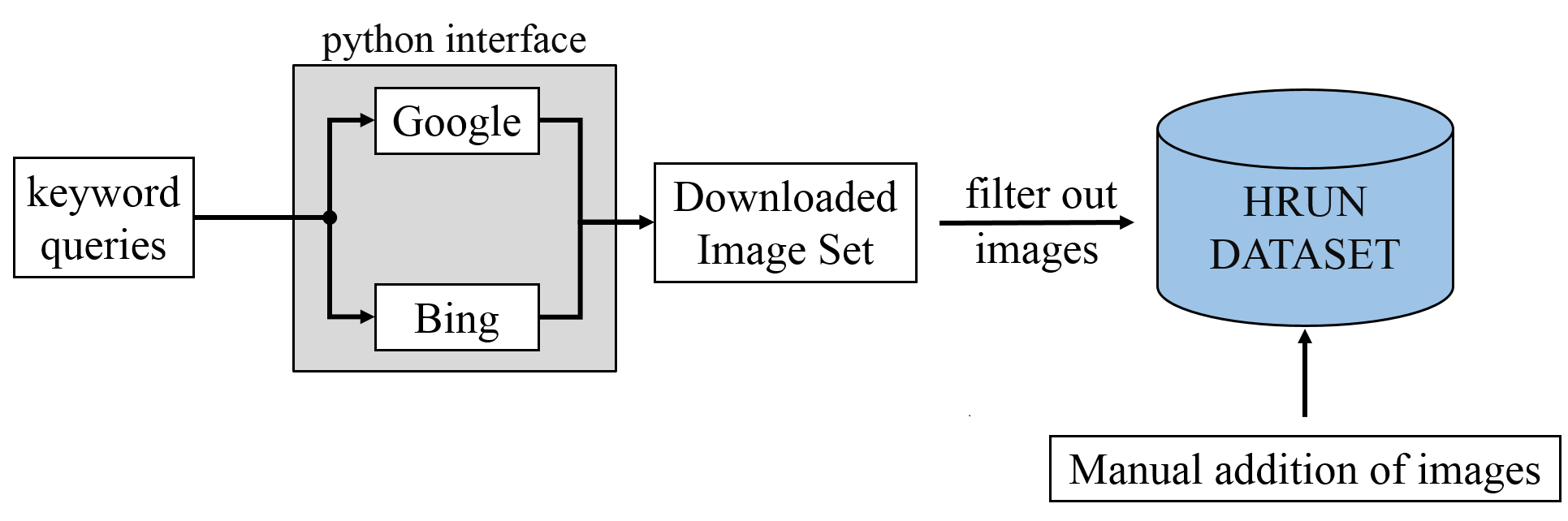}
\caption{Constructing HRUN Dataset}
\label{Constructing HRUN}
\end{figure}

This section describes the image collection procedure utilised for the formulation of the HRUN dataset, as captured by Figure \ref{Constructing HRUN}.

\begin{figure}[!t]
\centering
\includegraphics[height=5cm,width=7.5cm,keepaspectratio,center]{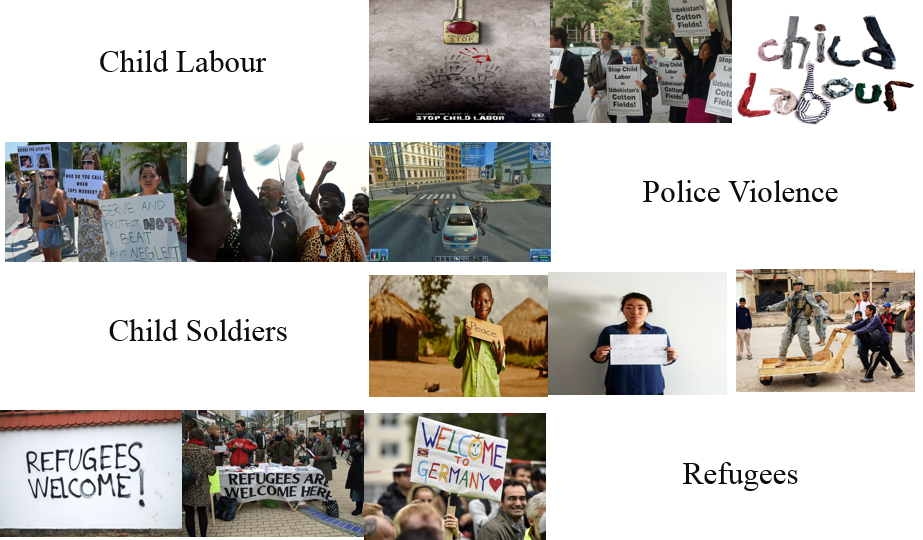}
\caption{Side by side examples of irrelevant images with their respective query term which were eliminated during the filtering process.}
\label{irrelevant images}
\end{figure}

\begin{table}[!b]
\centering
\caption{Image Collection Analysis from Search Engines}
\label{Collection Analysis}
\resizebox{\columnwidth}{!}{
\begin{tabular}{|c|cc|ccc|c|}
\hline
\rowcolor[HTML]{C0C0C0} 
\cellcolor[HTML]{C0C0C0}                                      & \multicolumn{2}{c|}{\cellcolor[HTML]{C0C0C0}\textbf{Retrieved Images}} & \multicolumn{3}{c|}{\cellcolor[HTML]{C0C0C0}\textbf{Relevant Images}} & \cellcolor[HTML]{C0C0C0}                                \\
\rowcolor[HTML]{C0C0C0} 
\multirow{-2}{*}{\cellcolor[HTML]{C0C0C0}\textbf{Query Term}} & \textbf{Google}                     & \textbf{Bing}                    & \textbf{Google}       & \textbf{Bing}       & \textbf{Manually}       & \multirow{-2}{*}{\cellcolor[HTML]{C0C0C0}\textbf{HRUN}} \\ \hline
Child labour                                                  & 99                                  & 137                              & 18                    & 5                   & 77                      & 100                                                     \\
Child Soldiers                                                & 176                                 & 159                              & 31                    & 13                  & 56                      & 100                                                     \\
Police Violence                                               & 149                                 & 232                              & 10                    & 16                  & 74                      & 100                                                     \\
Refugees                                                      & 111                                 & 140                              & 10                    & 39                  & 51                      & 100                                                     \\ \hline
\end{tabular}
}
\end{table}
Initially, the keywords, with a view to formulate the query terms, were collected in collaboration with specialists in the human rights domain. This happens in order to include more than one query term for every `targeted class'. For instance, for the class police violence the queries `police violence', `police brutality' and `police abuse of force' were all used for retrieving results. Work commenced with the Flickr photo-sharing website, but in a short time, it became apparent that its limitations resulted in a huge number of irrelevant results returned for the given queries. This happens because Flickr users are authorised to tag their uploaded images without restriction. Subsequently there have been situations where the given keyword was `armed conflict' and the majority of the returned images had to do with military parades. Another similar example was with the given keyword `genocide' where the returned results included protesting campaigns against genocide, something that may be consider close to the keyword, but it can not serve our purpose by any means. Another shortcoming was the case when people massively tagged an image deliberately incorrectly in order to acquire an increased number of hits on the photo-sharing website. Consequently, Google and Bing search engines were chosen as a better alternative. Images were downloaded for each class using a python interface to the Google and Bing application programming interfaces (APIs), with the maximum number of images permitted by their respective API for each query term. All exact duplicate images were eliminated from the downloaded image set, alongside images regarded as inappropriate during the filtering step as illustrated by Figure \ref{irrelevant images}. Nonetheless, the number of filtered images generated was still insufficient as shown in Table \ref{Collection Analysis}. 

For this reason, there were manually added other suitable images in order to reach the final structure of the HRUN dataset. We finally ended up with a total of four different categories, each one containing 100 distinct images of human rights violations captured in real world situations and surroundings. With this first attempt, our main intention was to produce a high quality dataset for the task in hand. For that reason, the number of categories was kept to a certain degree for the time being. Expanding the dataset both in categories and number of images has already been included in our actual future plans and many other online repositories that might be related to human rights violations are being checked into thoroughly.

\begin{figure*}[!t]
\includegraphics[height=5cm,width=\textwidth,keepaspectratio,center]{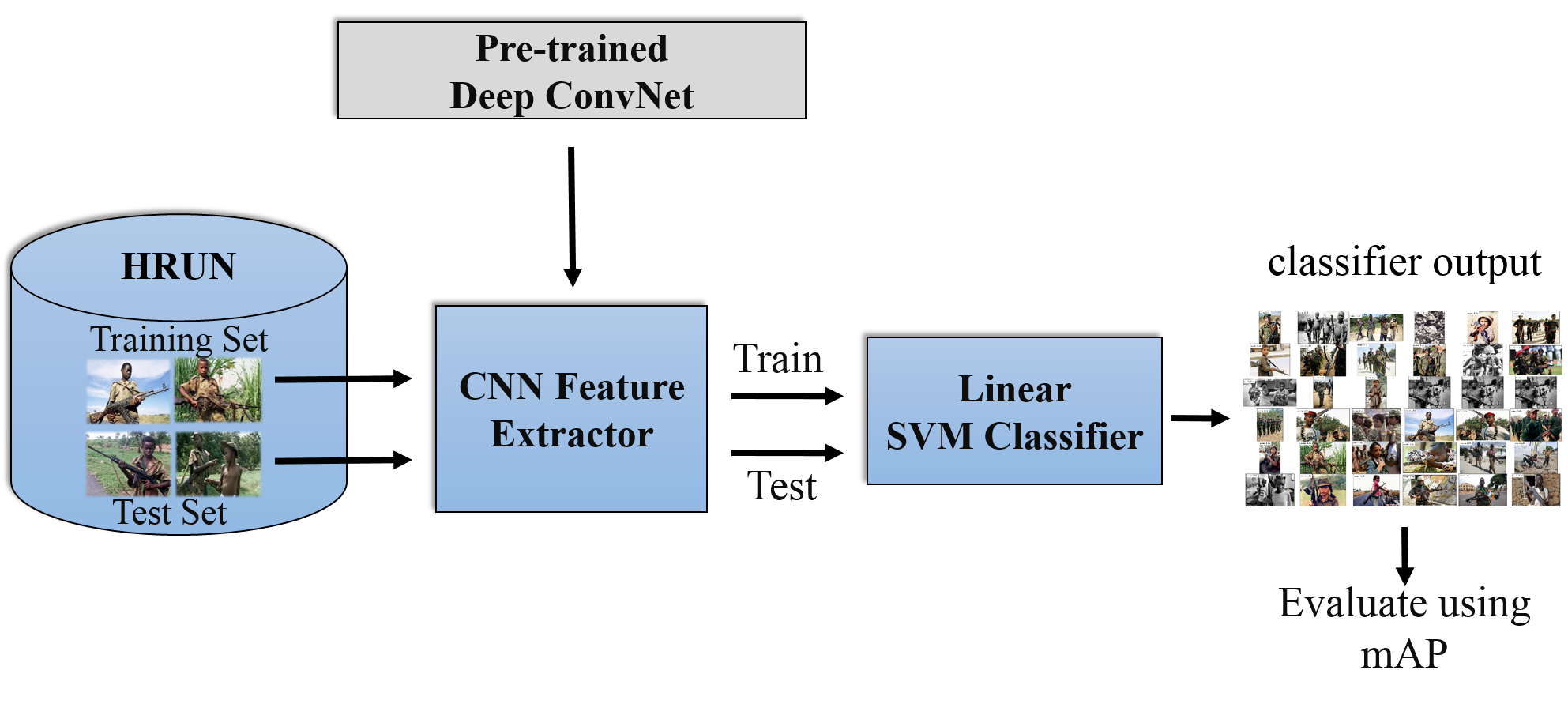}
\caption{An overview of the human rights violations
recognition pipeline used here. Different deep convolutional models are plugged into the pipeline one at a time, while the
training and test samples taken from the HRUN dataset remain fixed. Mean average
precision(mAP) metric is used for evaluating the results.}
\label{pipeline}
\end{figure*}

\section{\uppercase{Learning Deep Representations for Human Rights Violations Recognition}}

\subsection{Transfer Learning}
\noindent Our goal is to train a system that recognises different human rights violations from a given input image of the HRUN dataset. One high-priority research issue in our work is how to find a good representation for instances in such a unique domain. More than that, having a dataset of sufficient size is problematic for this task as described in the previous section. For those two reasons, a conventional alternative to training a deep ConvNet from the very beginning, is to use a pre-trained model and then use the ConvNet as a fixed feature extractor for the task of interest. This method, referred to as \textit{transfer learning}\cite{17,32}, is implemented by taking a pre-trained CNN, replacing the fully-connected layers (and potentially the last convolutional layer), and consider the rest of the ConvNet as a fixed feature extractor for the relevant dataset. By freezing the weights of the convolutional layers, the deep ConvNet can still extract general image features such as edges, while the fully connected layers can take this information and use it to classify the data in a way that is applicable to the problem.

\subsection{Pipeline for Human Rights Violations Recognition}
\noindent The entire pipeline used for the experiments is depicted in Figure \ref{pipeline}, and detailed further below. In this pipeline, every block is fixed except the feature extractor as different deep convolutional networks are plugged in, one at a time, to compare their performance utilizing the mean average precision (mAP) metric. 

Given a training dataset $T_{r}$ consisting of m human rights violation categories, a test dataset $T_{s}$ comprising unseen images of the categories given in $T_{r}$, and a set of n pre-trained CNN architectures ($C_{1}$,...$C_{n}$), the pipeline operates as follows: The training dataset $T_{r}$ is used as input to the first CNN architecture $C_{1}$. The output of $C_{1}$, as described above,  is then utilized to train m SVM classifiers. Once trained, the test dataset $T_{s}$ is employed to assess the performance of the pipeline using mAP. The training and testing procedures are then repeated after replacing $C_{1}$ with the second CNN architecture $C_{2}$ to evaluate the performance of the human rights violation recognition pipeline. For a set of n pre-trained CNN architectures, the training and testing processes are repeated n times. Since the entire pipeline is fixed (including the training and test datasets, learning procedure and evaluation protocol) for all n CNN architectures, the differences in the performance of the classification pipeline can be attributed to the specific CNN architectures used.
For comparison, 10 different deep CNN architectures were identified, grouped by the common paper which they were first made public: a) 50-layer ResNet, 101-layer ResNet and 152-layer ResNet presented in \cite{11}; b) 22-layer GoogLeNet \cite{12}; c) 16-layer VGG-Net and 19-layer VGG-Net introduced in \cite{10} ; d) 8-layer VGG-S, 8-layer VGG-M and 8-layer VGG-F displayed in \cite{14}; and e) 8-layer Places \cite{16}, as they represent the state-of-the-art for image classification tasks. For further design and implementation details for these models, please refer to their respective papers. 
To ensure a fair comparison, all the standardised CNN models used in our experiments are based on the opensource Caffe framework \cite{31} and are pre-trained on 1000 ImageNet \cite{8} classes with the exception of Places CNN \cite{16} which was trained on 205 scenes categories of Places database. For the majority of the networks, the dimensionality of the last hidden layer (FC7) leads to a 4096x1 dimensional  image representation. Since the GoogLeNet \cite{12} and the ResNet \cite{11} architectures do not utilise fully connected layers at the end of their networks, the last hidden layers before average pooling at the top of the ConvNet are exploited with 1024x7x7 and 2048x7x7 feature maps respectively, to counterbalance the behaviour of the pool layers, which provide downsampling regarding the spatial dimensions of the input.

\begin{table*}[!t]
\centering
\caption{\textbf{Human rights violations classification results with a 70/30 split for training and testing images}. Mean average precision (mAP) accuracy for different CNNs. Bold font highlights the leading mAP result for every experiment.}
\label{30/70}
\resizebox{\textwidth}{!}{
\begin{tabular}{ccccccc}
\rowcolor[HTML]{C0C0C0} 
\cellcolor[HTML]{C0C0C0}                                 & \textbf{Dimensional}    & \cellcolor[HTML]{C0C0C0}                               & \cellcolor[HTML]{C0C0C0}                                        & \cellcolor[HTML]{C0C0C0}                                          & \cellcolor[HTML]{C0C0C0}                                           & \cellcolor[HTML]{C0C0C0}                                    \\
\rowcolor[HTML]{C0C0C0} 
\multirow{-2}{*}{\cellcolor[HTML]{C0C0C0}\textbf{Model}} & \textbf{Representation} & \multirow{-2}{*}{\cellcolor[HTML]{C0C0C0}\textbf{mAP}} & \multirow{-2}{*}{\cellcolor[HTML]{C0C0C0}\textbf{Child Labour}} & \multirow{-2}{*}{\cellcolor[HTML]{C0C0C0}\textbf{Child Soldiers}} & \multirow{-2}{*}{\cellcolor[HTML]{C0C0C0}\textbf{Police Violence}} & \multirow{-2}{*}{\cellcolor[HTML]{C0C0C0}\textbf{Refugees}} \\
ResNet 50                                                & 100K                     & 42.59                                                  & 41.12                                                           & 43.69                                                             & 43.81                                                              & 41.73                                                       \\
ResNet 101                                               & 100K                     & 42.07                                                  & 40.48                                                           & 44.78                                                             & 42.56                                                              & 40.48                                                       \\
ResNet 152                                               & 100K                     & 45.80                                                  & 44.27                                                           & 44.11                                                             & 48.08                                                              & 46.73                                                       \\ \hline
GoogLeNet                                                & 50K                      & 48.62                                                  & 42.72                                                           & 40.71                                                             & 61.91                                                              & 49.16                                                       \\ \hline
VGG 16                                                   & 4K                      & 77.46                                                  & 70.79                                                           & \textbf{77.71}                                                    & 83.46                                                              & 77.87                                                       \\
VGG 19                                                   & 4K                      & 47.01                                                  & 31.69                                                           & 50.98                                                             & 73.79                                                              & 31.57                                                       \\ \hline
VGG - M                                                  & 4K                      & 67.93                                                  & 59.52                                                           & 62.96                                                             & 81.45                                                              & 67.80                                                       \\
VGG - S                                                  & 4K                      & \textbf{78.19}                                         & \textbf{80.17}                                                  & 64.46                                                             & 87.46                                                              & \textbf{80.68}                                              \\
VGG - F                                                  & 4K                      & 64.15                                                  & 45.42                                                           & 63.20                                                             & 84.78                                                              & 63.21                                                       \\ \hline
Places                                                   & 4K                      & 68.59                                                  & 55.67                                                           & 65.60                                                             & \textbf{93.17}                                                     & 59.92                                                      
\end{tabular}}
\end{table*}

\begin{table*}[!t]
\centering
\caption{\textbf{Human rights violations classification results with a 50/50 split for training and testing images}. Mean average precision (mAP) accuracy for different CNNs. Bold font highlights the leading mAP result for every experiment.}
\label{50/50}
\resizebox{\textwidth}{!}{
\begin{tabular}{ccccccc}
\rowcolor[HTML]{C0C0C0} 
\cellcolor[HTML]{C0C0C0}                                 & \textbf{Dimensional}    & \cellcolor[HTML]{C0C0C0}                               & \cellcolor[HTML]{C0C0C0}                                        & \cellcolor[HTML]{C0C0C0}                                          & \cellcolor[HTML]{C0C0C0}                                           & \cellcolor[HTML]{C0C0C0}                                    \\
\rowcolor[HTML]{C0C0C0} 
\multirow{-2}{*}{\cellcolor[HTML]{C0C0C0}\textbf{Model}} & \textbf{Representation} & \multirow{-2}{*}{\cellcolor[HTML]{C0C0C0}\textbf{mAP}} & \multirow{-2}{*}{\cellcolor[HTML]{C0C0C0}\textbf{Child Labour}} & \multirow{-2}{*}{\cellcolor[HTML]{C0C0C0}\textbf{Child Soldiers}} & \multirow{-2}{*}{\cellcolor[HTML]{C0C0C0}\textbf{Police Violence}} & \multirow{-2}{*}{\cellcolor[HTML]{C0C0C0}\textbf{Refugees}} \\
ResNet 50                                                & 100K                     & 70.94                                                  & 73.15                                                           & 68.07                                                             & 70.44                                                              & 72.09                                                       \\
ResNet 101                                               & 100K                     & 68.46                                                  & 69.50                                                           & 66.90                                                             & 68.34                                                              & 69.09                                                       \\
ResNet 152                                               & 100K                     & 76.20                                                  & 80.60                                                           & 73.07                                                             & 72.00                                                              & 79.12                                                       \\ \hline
GoogLeNet                                                & 50K                      & 55.92                                                  & 41.48                                                           & 60.21                                                             & 55.52                                                              & 66.48                                                       \\ \hline
VGG 16                                                   & 4K                      & 84.79                                                  & 79.15                                                           & 87.94                                                             & 89.47                                                              & 82.59                                                       \\
VGG 19                                                   & 4K                      & 60.39                                                  & 35.72                                                           & 72.67                                                             & 83.10                                                              & 50.08                                                       \\ \hline
VGG - M                                                  & 4K                      & 78.94                                                  & 68.71                                                           & 82.32                                                             & 89.99                                                              & 74.74                                                       \\
VGG - S                                                  & 4K                      & \textbf{88.10}                                         & \textbf{84.84}                                                  & 88.14                                                             & 91.92                                                              & \textbf{87.50}                                              \\
VGG - F                                                  & 4K                      & 73.46                                                  & 53.57                                                           & 78.78                                                             & 90.41                                                              & 71.08                                                       \\ \hline
Places                                                   & 4K                      & 81.40                                                  & 62.04                                                           & \textbf{89.97}                                                    & \textbf{95.70}                                                     & 77.90                                                      
\end{tabular}}
\end{table*}

\section{\uppercase{Experiments and Results}}

\subsection{Evaluation Details}
\noindent The evaluation process is divided into two different sets of scenarios, each one making use of an explicit split of images between the training and testing samples of the pipeline. For the first scenario, a split of 70/30 was utilised, while for the second scenario the split was adjusted to 50/50 for training and testing images respectively. Additionally, three distinct series of tests were conducted for each scenario, each and every one assembled with a completely  arbitrary shift of the entire image set for every category of the HRUN dataset. This approach ensures an unbiased comparison with a rather limited dataset like HRUN at present. The compound results of all three tests are given in Table \ref{30/70} and Table \ref{50/50} and analysed below.

\subsection{Results and Discussion}
\noindent It is evident from Table \ref{30/70} and Table \ref{50/50} that the \textit{Slow} CNN architecture performs the best for the child labour category for both scenarios. \textit{VGG} with 16 layers performs the best in the case of child soldiers with scenario 1, while on the other hand, scenario's 2 best performing architecture is \textit{Places} with \textit{VGG-16} coming genuinely close. Places was also the best performing architecture for the category of police violence for the two scenarios. Lastly, regarding refugees category, the Slow version of \textit{VGG} was the dominant architecture for both scenarios.
Since our work is the first effort in the literature to recognise human rights violations, we are not able to compare our experimental results with other works. 

\begin{figure*}[!t]
\centering
\includegraphics[height=19cm,width=\textwidth,keepaspectratio,center]{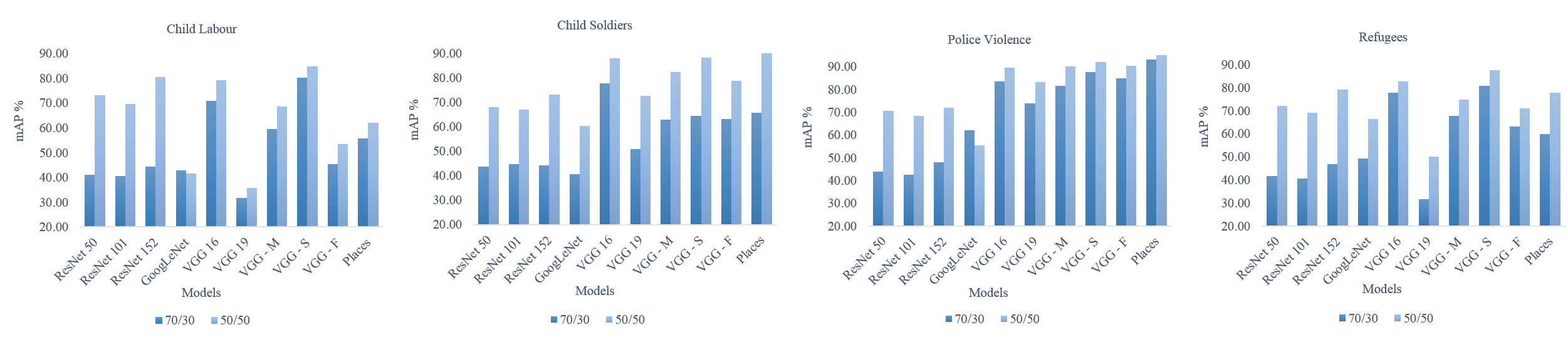}
\caption{Comparison of deep convolutional networks performance, with reference to mAP, for the two diverse scenarios appearing in our experiments. The number on the left side of the slash denotes the training proportion of images,
while the name on the right implies the testing percentage.}
\label{mAP comparison}
\end{figure*}
\noindent However the results are unquestionably promising and reveal that the best performing CNN architectures can achieve up to 88.10\% mean average precision when recognising human rights violations. On the other hand, some of the regularly top performing deep ConvNets, such as \textit{GoogLeNet} and \textit{ResNet}, fell short for this particular task compared to the others. Such weaker performance occurs primarily because of the limited dataset size, whereby learning millions of parameters of those very deep convolutional networks is usually impractical and may lead to over-fitting. Another interpretation could be due to the inadequate structure of the image representation deducted from the last hidden layer before average pooling compared to the FC7 layer of the others.   
Furthermore, it is clear that by utilising the 50/50 split of images in the course of scenario 2, there is a considerable boost in performance of the human rights violations recognition pipeline as compared to the first scenario when a split of 70/30 was employed for training and testing images respectively. Figure \ref{mAP comparison} depicts the effect of two varying training data sizes (scenario 1 vs scenario 2) on the performance of different deep convolutional networks. Remarkably with scenario 2, where the half and half split was applied, accomplishes a notable improvement on mean average precision which spans from 4.03\% up to 36.33\% across all four HRUN categories which were tested. Only on two occassions scenario 2 was outperformed by scenario 1, both of them while \textit{GoogLeNet} was selected for the categories of `child labour' and `police violence'. This observation strengthens the point of view discussed above relative to the last hidden layer of this model. Nonetheless, in all instances a mean average precision greatly above 40\% was achieved, which can be regarded as an impressive outcome given the unconventional nature of the problem, the limited dataset which was adopted for learning deep representations and the transfer learning approach that was employed here.

\section{\uppercase{Conclusions}}
\noindent Recognising human rights violations through digital images is a new and challenging problem in the area of computer vision. We introduce a new, open human rights understanding dataset, HRUN, designed to represent human rights and international humanitarian law violations found in the real world. Using this innovative dataset we conduct an evaluation of recent deep learning architectures for human rights violations recognition and achieve results that are comparable to prior attempts on other long-standing hallmark tasks of computer vision in the hope that it would provide a scaffold for future evaluations, and good benchmark for human rights advocacy research. The following conclusions have derived:
Digital images that can be rated as appropriate for human rights monitoring purposes are rare and characterising them requires great effort, expertise and vast time.
Utilising transfer learning for the task of recognising human rights violations can provide very strong results by employing a straightforward combination of deep representations and a linear SVM.
Deep convolutional neural networks are constructed to benefit and learn from massive amounts of data. For this reason and in order to obtain even higher quality recognition results, training a deep convolutional network from scratch on an expanded version of the HRUN dataset is likely to further improve results.
Inspired by the high-standard characteristics of legal evidence, in the future we would like to have the means to clarify three different questions set by every human rights monitoring mechanism: \textit{what}, \textit{who} and \textit{how}, and expand our dataset to a wider range of categories in order to include them. We also presume that further analysis of joint object recognition and scene understanding will be beneficial and lead to improvements in both tasks for human rights violations understanding. 

\section*{\uppercase{Acknowledgements}}
\noindent We acknowledge MoD/Dstl and EPSRC for providing the grant to support the UK academics (Ales Leonardis) involvement in a Department of Defense funded MURI project. This work was also supported in part by EU H2020 RoMaNS 645582, EPSRC EP/M026477/1 and ES/M010236/1.


\bibliographystyle{apalike}
{\small
\bibliography{example}}

\end{document}